\documentclass[csla]{academic}
\usepackage{amsmath}
\usepackage{epsfig}
\usepackage{natbib}    
\bibpunct[,]{(}{)}{;}{a}{,}{,}
\let\cbstart\relax
\let\cbend\relax

\begin{document}

\shortauthor{L. Mangu {\em et al.}}
\shorttitle{Finding Consensus in Speech Recognition}

\title{Finding consensus in speech recognition:\\
word error minimization and other applications of confusion networks%
\thanks{Some of the material presented here was previously published in
conference proceedings \protect\citep{Mangu:euro99,ManguBrill:99}.}
}

\author[1]{Lidia Mangu\thanks{\email{mangu@us.ibm.com}}%
\thanks{Correspondence to: L. Mangu}}
\address[1]{IBM Watson Research Center,
PO Box 704, Yorktown Heights, NY 10598, U.S.A.,}
\author[2]{Eric Brill\thanks{\email{brill@microsoft.com}}}
\address[2]{Microsoft Research, One Microsoft Way, Redmond, WA 98052, U.S.A.,}
\author[3]{Andreas Stolcke\thanks{\email{stolcke@speech.sri.com}}}
\address[3]{SRI International, 333 Ravenswood Ave., Menlo Park, CA 94025, U.S.A.}


\maketitle

\renewcommand\thefootnote{\arabic{footnote}}

\begin{abstract}%
We describe a new framework for distilling information from word
lattices to improve the accuracy of speech recognition and obtain a 
more perspicuous representation of a set of alternative hypotheses.
In the standard MAP decoding approach the recognizer outputs the string of 
words corresponding to the path with the highest posterior probability
given the acoustics and a language model. However, even given optimal 
models, the MAP decoder does not necessarily minimize the commonly used
performance metric, word error rate (WER). We describe a method 
for explicitly minimizing WER by extracting word hypotheses with the 
highest posterior probabilities from word lattices. We change the 
standard problem formulation by replacing global search over a large 
set of sentence hypotheses with local search over a small set of word 
candidates. In addition to improving the accuracy of the recognizer, 
our method produces a new representation of the set of 
candidate hypotheses that specifies the sequence of word-level confusions in 
a compact lattice format.
We study the properties of confusion networks and examine their
use for other tasks, such as lattice compression, word spotting,
confidence annotation, and reevaluation of recognition
hypotheses using higher-level knowledge sources.
\end{abstract}


\renewcommand{\.}{$\cdot$}
 
\section{Introduction}
\label{sec:intro}
In the standard maximum {\em a posteriori} probability (MAP) decoding
approach to speech recognition \citep*{Bahl:83}
the recognizer outputs  the string of word hypotheses
corresponding to the path with  the highest posterior probability given the
acoustics and a language model.  However, even given optimal models, the MAP
decoder does not necessarily minimize the commonly used performance criterion
for recognition, the word error rate (WER). Intuitively, one should 
maximize {\em word} posterior probabilities instead of whole
sentence posteriors to minimize WER.

Prior work \citep*{StolckeEtAl:eurospeech97} has shown how WER can be
explicitly minimized in an $N$-best rescoring algorithm.
That approach is suboptimal because it restricts hypothesis choice to
a rather small set compared to the search space of the recognizer.
In this paper we present a new word error minimization algorithm that
is applicable to word lattices, or partially ordered networks of 
word hypotheses.  
Word lattices are used by many large vocabulary recognizers as a compact
intermediate representation of alternative hypotheses and contain 
orders of magnitude more hypotheses than typical $N$-best lists.
Word error minimization applied to 
word lattices promises better performance than $N$-best lists for two 
reasons.
First, lattices provide a larger set of hypotheses from which to choose;
second, the more accurate representation of the hypothesis space gives 
better estimates for word posterior probabilities and, consequently, of
expected word error.
However, as we will see below, the lattice representation also leads to new
computational problems: it is no longer feasible to compute word errors
between hypotheses exactly.
The solution will be to minimize a slightly modified 
word error function that can be computed efficiently
and is empirically shown to closely approximate the standard word error.

In Section~\ref{sec:approach} we motivate our approach on theoretical
and empirical grounds, and survey prior work.
Section~\ref{sec:algorithm} describes the algorithm itself.
Section~\ref{sec:results} gives an experimental evaluation of our
method in terms of recognition accuracy, followed by further diagnostic 
experiments in Section~\ref{sec:analysis}.
In Section~\ref{sec:apl_conf_nets} we describe various
properties of {\em confusion networks}, a specialized form of lattice generated
by our method, and show how they can be employed for tasks other than 
word recognition.
Section~\ref{sec:other} compares our approach to a recently developed
alternative method for lattice-based word error minimization.
Section~\ref{sec:conclusion} concludes.

\newcommand{\WE}{{\it WE}}

\section{Motivation and prior work}
        \label{sec:approach}

\subsection{Theoretical motivation}
\label{sec:wemin}
We can motivate our approach
from a theoretical point of view by the mismatch between the standard scoring paradigm (MAP) and the commonly used performance metric (WER). 
In the standard approach to speech recognition \citep{Bahl:83}, the 
goal is to find the sentence hypothesis that maximizes the posterior
probability $P(W | A)$ of the word sequence $W$ given the
acoustic information $A$.  We call this the {\em sentence MAP} approach.
Sentence posteriors are then usually approximated as the product of a 
number of knowledge sources, and normalized.  For example, given
a language model $P(W)$ and acoustic likelihoods $P(A|W)$, we can approximate%
\footnote{The normalization can be omitted for purposes of posterior maximization, 
but is made explicit here for clarity.}
\begin{equation}
        \label{eq:posterior}
        P(W|A) \approx { \frac{P(W) P(A|W)}
                        {\sum_{k} P(W^{(k)}) P(A|W^{(k)})} }
\end{equation}
where $k$ ranges over the set of hypotheses generated by the recognizer. 

Bayesian decision theory [e.g.\ \citet*{Duda:73}] tells us that maximizing
sentence posteriors minimizes the {\em sentence level error} (the probability of having
at least one error in the sentence string). 
However, the commonly used performance metric in speech recognition is
{\em word error},
i.e.\ the Levenshtein (string edit) distance $\WE(W,R)$ between
a hypothesis $W$ and the reference string $R$. $\WE(W,R)$ is defined as
the number of substitutions, deletions, and insertions in $W$ relative to $R$ under an alignment of the two strings that minimizes a weighted combination of these error types.  The string edit distance is a more forgiving
(and for many applications more relevant) error metric because it
gives partial credit for correctly recognized portions of sentences.

Sentence error and word error rates are assumed to be highly correlated,
so minimizing one would tend to minimize the other. However, as our 
empirical results will show, there is a significant difference between
optimizing for sentence vs.\ word error rate.
To gain an intuitive understanding of this difference
it is helpful to examine an example.

\begin{table}
\caption{Example illustrating the difference between sentence and word error measures.}
\label{tab:example}
\begin{center}
\begin{tabular}{cccccccc}
\hline
\multicolumn{3}{c}{Hypothesis ($H$)} & & & & & \\
\cline{1-3}
$w_1$ & $w_2$ & $w_3$ & $P(H | A)$ & $P(w_1|A)$ & $P(w_2|A)$ & $P(w_3|A)$ & $E[\mbox{correct}]$ \\
\hline
I & DO & INSIDE  & 0\.16           &  0\.34         & 0\.29      & 0\.16 &  0\.79\\
I & DO & FINE    & 0\.13           &  0\.34         & 0\.29      & 0\.28  & 0\.91\\
BY & DOING & FINE & 0\.11          &  0\.45         & 0\.49      & 0\.28  & 1\.22\\ 
BY & DOING & WELL & 0\.11          &  0\.45         & 0\.49      & 0\.11  & 1\.05\\
BY & DOING & SIGHT & 0\.10         &  0\.45         & 0\.49      & 0\.10  & 1\.04\\
BY & DOING & BYE & 0\.07           &  0\.45         & 0\.49      & 0\.07  & 1\.01\\
BY & DOING & THOUGHT & 0\.05       &  0\.45         & 0\.49      & 0\.07  & 0\.99\\
I & DOING & FINE & 0\.04           &  0\.34         & 0\.49      & 0\.28  & 1\.11\\
I & DON'T & BUY & 0\.01            &  0\.34         & 0\.01      & 0\.01  & 0\.36\\
BY & DOING & FUN & 0\.01           &  0\.45         & 0\.49      & 0\.01  & 0\.95\\
\hline
\end{tabular}
\end{center}
\end{table}

The first column in Table~\ref{tab:example} shows a 10-best list of
hypotheses that are produced by the recognizer, and the second column shows the
corresponding joint posterior probabilities $P(H|A)$ for these
hypotheses.  Columns 3, 4 and 5 give the posterior probabilities
$P(w|A)$ for individual words. These posterior word probabilities
follow from the joint posterior probabilities by summing over all
hypotheses that share a word in a given position. Column 6 shows the
expected number of correct words $E[\mbox{correct}]$ in each
hypothesis, under the posterior distribution. This is simply the sum of
the individual word posterior probabilities, since
\begin{eqnarray}
&&      E[ \mbox{words correct}(w_1 w_2 w_3) | A ] \nonumber \\
&& \quad = E[ \mbox{correct}(w_1) | A ] + E[ \mbox{correct}(w_2) | A ] + E[ \mbox{correct}(w_3) | A ]\\
&& \quad = P( w_1 | A ) + P(w_2 | A ) + P(w_3 | A ) \nonumber
\end{eqnarray}
As can be seen, although the hypothesis ``BY DOING FINE'' does not have
the highest posterior, it has the highest expected number of correct
words, i.e.\ the minimum expected word error. The correct answer for
this example is ``I'M DOING FINE'', which means that the MAP hypothesis
``I DO INSIDE'' has misrecognized all the words ($WER = 3$), whereas
the new hypothesis recognized incorrectly only one word ($WER = 1$).
Thus, we have shown that optimizing overall posterior probability
(sentence error) does not always minimize expected word error.  This
happened because words with high posterior probability did not have
high posterior probability when combined.

\subsection{Word error minimization}

Given word error as our objective function, we can replace the MAP approach
with a new hypothesis selection approach based on minimizing the expected
word error under the posterior distribution:
\begin{equation}
        \label{eq:ewe}
         E_{P(R|A)} [\WE(W,R)] = \sum_{R} P(R|A) \WE(W,R)
\end{equation}
This equation provides a general recipe for computing expected 
word-level error from sentence-level posterior estimates.

A direct algorithmic version involves two iterations: a summation over
potential references $R$ and a minimization over hypotheses $W$.
Therefore, the amount of work grows quadratically with the number of 
hypotheses.  Furthermore, the computation of the word error function
is nontrivial: it involves a dynamic programming alignment of 
$W$ and $R$ and takes time proportional to the square of the hypotheses 
lengths.
This raises the question of whether explicit word error minimization can be
carried out feasibly for large vocabulary recognition.
The next section briefly reviews a prior approach based on an
$N$-best approximation.  The remainder of the paper is devoted to
an improved approach considering the hypotheses in a word lattice.

\subsection{An $N$-best approximation}
	\label{sec:nbest}
\newcommand{\argmax}{\mathop{\rm argmax}}
\newcommand{\argmin}{\mathop{\rm argmin}}

In previous work \citep{StolckeEtAl:eurospeech97},
word error minimization was implemented by limiting 
the search space and the expected word error computation to an
$N$-best list, i.e.
letting both $W$ and $R$ range over the $N$ best hypothesis output by a recognizer:
\begin{equation}
        \label{eq:center}
        W_c = \argmin_{i=1,N} \sum_{k=1}^{N} P(R^{(k)}|A) \WE(W^{(i)}, R^{(k)})
\end{equation}
We refer to the hypothesis $W_c$ thus obtained as the {\em center hypothesis}.
An optimized algorithm that finds the center hypothesis and scales linearly
with $N$ in practice is given in Appendix~\ref{app:nbest-alg}.

\cbstart
\citet*{Goel:98} have noted that the brute-force optimization of 
Equation~(\ref{eq:ewe}) and its $N$-best approximation (\ref{eq:center})
generalizes to
objective functions other than word error.
For example, they show \citep*{Goel:CSL} that the named-entity tagging
\cbend
performance on automatically recognized speech can be improved by
explicitly optimizing an appropriate tagging score (e.g.\ F-score).

\subsection{Lattice-based word error minimization}
\label{sec:lat}
Lattices represent a combinatorial number of sentence hypotheses,
presenting the potential to improve the $N$-best approach through both
more accurate error estimates [the summation in
Equation~(\ref{eq:ewe})] and a larger search space for minimization.

From a practical point of view, lattices are often generated as 
a preliminary step to $N$-best lists in a multipass recognition
system, and are thus obtained with less computational overhead.
Furthermore, efficient lattice generation techniques that are essentially no
more burdensome than simple 1-best recognition have recently been developed
\citep*{LjoljeEtAl:99}.
Therefore, being able to carry out word error minimization directly 
from lattices also represents a practical simplification and efficiency
enhancement of the overall recognition system.

In moving to lattice-based hypothesis selection, we are faced with 
a computational problem.  The number of hypotheses contained in a 
lattice is several orders of magnitude larger than in $N$-best lists of
practical size,
making a straightforward computation of the center hypothesis as
in Equation~(\ref{eq:center}) infeasible.
A natural approach to this problem is to exploit the structure of 
the lattice for efficient computation of the center hypothesis.
Unfortunately, there seems to be no efficient (e.g.\ dynamic programming)
algorithm of this kind.  The fundamental difficulty is that the objective
function is based on pairwise string distance, a nonlocal measure.
A single word difference anywhere in a lattice path can have global
consequences on the alignment of that path to other paths, preventing
a decomposition of the objective function that exploits the lattice
structure.

\newcommand{\MWE}{{\it MWE}}
To work around this problem, we decided to replace the original
pairwise string alignment (which gives rise to the standard string edit
distance $\WE(W,R)$) with a modified, multiple string alignment.
The new approach incorporates all lattice hypotheses\footnote{In practice
we apply some pruning of the lattice to remove low probability word
hypotheses (see Section~\ref{sec:pruning}).}
into a single alignment, and word error between any two hypotheses is then computed 
according to that one alignment. The multiple alignment thus defines 
a new string edit distance, which we will call $\MWE(W,R)$.
While the new alignment may in some cases overestimate the word error
between two hypotheses, as we will show in Section~\ref{sec:analysis}
it gives very similar results in practice.

\begin{figure}
\begin{itemize}
\item[(a)] Input lattice (``SIL'' marks pauses)
\begin{center}
\includegraphics[width=0.8\textwidth]{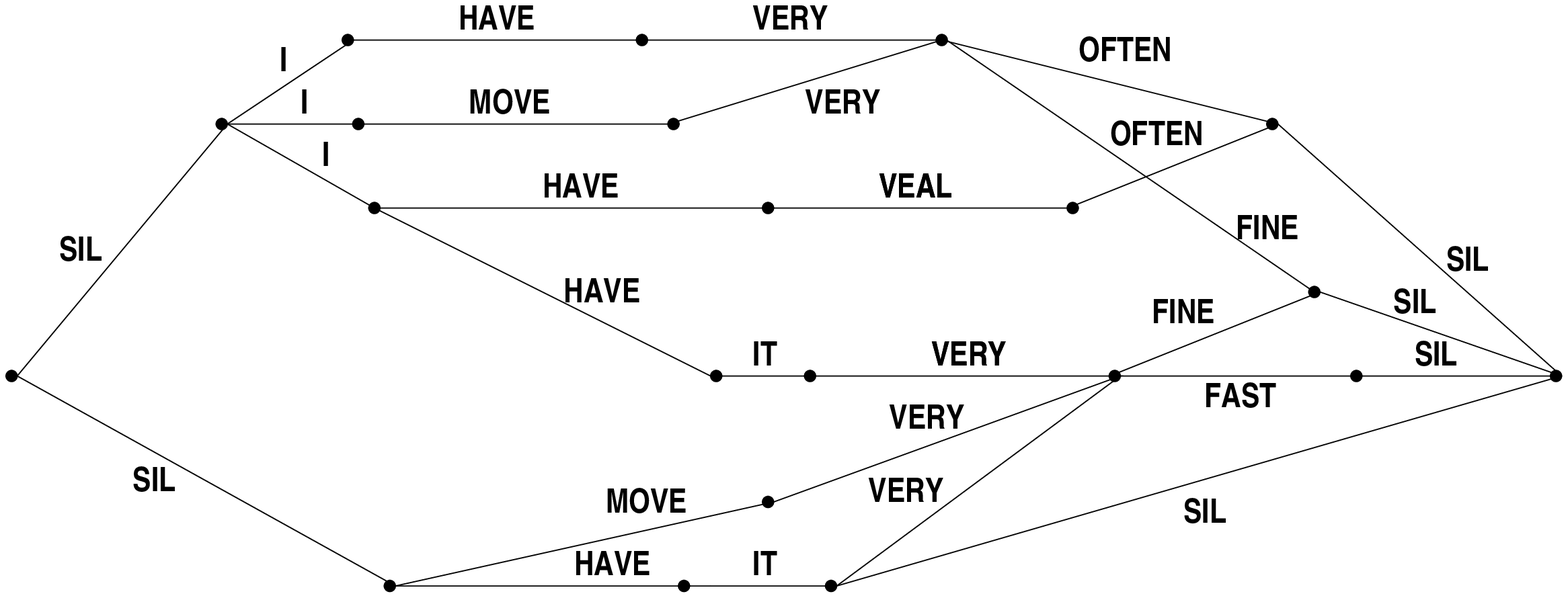}
\end{center}
\item[(b)] Multiple alignment (``-'' marks deletions)
\begin{center}
\includegraphics[width=0.6\textwidth]{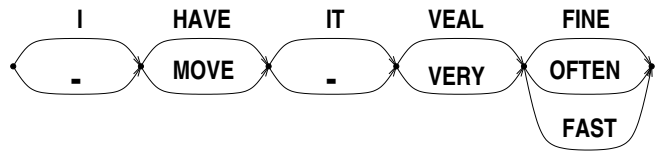}
\end{center}
\end{itemize}
\caption{Sample recognition lattice and corresponding multiple alignment
		represented as confusion network.}
        \label{fig:example}
\end{figure}

The main benefit of the multiple alignment is that it allows us to
extract the hypothesis with the smallest expected (modified) 
word error very efficiently.
To see this, consider an example. Figure~\ref{fig:example} shows a
word lattice and the corresponding hypothesis alignment.
Each word hypothesis is mapped to a position in the alignment
(with deletions marked by ``-'').
The alignment also supports the computation of {\em word posterior
probabilities}.  The posterior probability of a word hypothesis 
is the sum of the posterior probabilities of all lattice paths of which
the word is a part.
Given an alignment and posterior probabilities, it is easy to see
that the hypothesis with the lowest expected word error is obtained
by picking the word with the highest posterior at each position in
the alignment.
We call this the {\em consensus hypothesis}.

\section{The algorithm}
        \label{sec:algorithm}

Having given an intuitive idea of word error minimization based on 
lattice alignment, we can now make these notions more precise and 
describe the algorithm in detail. 

Pseudo-code for the crucial steps and information on algorithmic complexity
is given in Appendix~\ref{app:lattice-alg}.

\subsection{Lattice alignment}
\label{sec:align}

As we saw, the main complexity of the approach is in finding 
a good multiple alignment of lattice hypotheses, i.e.\ one that
approximates the pairwise alignments.
Once an alignment is found we can determine the minimizing
word hypothesis exactly. 
Finding the optimal alignment itself is a problem for which no
efficient solution is known \citep*{Gusfield:92}. 
We therefore resort to a heuristic approach based on lattice topology, 
as well as time and phonetic information associated with word hypotheses.

\newcommand{\Inode}{{\it Inode}}
\newcommand{\Fnode}{{\it Fnode}}
\newcommand{\Itime}{{\it Itime}}
\newcommand{\Ftime}{{\it Ftime}}
\newcommand{\Word}{{\it Word}}
\newcommand{\Words}{{\it Words}}
Let $E$ be the set of links (or edges) in a word lattice,
each link $e$ being characterized by its starting node
$\Inode(e)$, ending node $\Fnode(e)$, starting time $\Itime(e)$,
ending time $\Ftime(e)$, and word label $\Word(e)$.
From the acoustic and language model scores in the lattice, we can also
compute the posterior probability $p(e)$ of each link, i.e.\ the sum
of posteriors of all paths through $e$.  Furthermore, for a given
link subset $F \subset E$, let 
\[
        \Words(F) = \{ w | \exists e \in F: \Word(e) = w \}
\]
be its set of words, and
\[
        p(F) = \sum_{e \in F} p(e)
\]
its total posterior probability.

Formally, an alignment consists of an equivalence relation over the word
hypotheses (edges) in the lattice, together with a total ordering of the 
equivalence classes, such that the ordering is consistent with that of the 
original lattice.  Each equivalence class corresponds to one ``position''
in the alignment, and the members of a class are those word hypotheses that
are ``aligned to each other,'' i.e.\ represent alternatives.
We use $[e]$ to denote the equivalence class of which $e$ is a member.

The lattice defines a partial order $\leq$ on the links.
For $e,f \in E$, $e \leq f$ iff
\begin{itemize}
\item $e = f$ or
\item $\Fnode(e) = \Inode(f)$ or
\item $\exists e' \in E$ such that $e \leq e'$ and $e' \leq f$ .
\end{itemize}
Informally, $e \leq f$ means that $e$ ``comes before'' $f$ in the lattice.

Now let ${\mathcal E} \subset {2^E}$ be a set of equivalence classes on $E$,
and let $\preceq$ be a partial order on $\mathcal E$. 
We say that $\preceq$ is {\em consistent} with the lattice order $\leq$
if $e_1 \leq e_2$ implies $[e_1] \preceq [e_2]$, for all
$e_1 \in [e_1], e_2 \in [e_2]$, $[e_1], [e_2] \in {\mathcal E}$.
Consistency means that the equivalence relation preserves the temporal
order of word hypotheses in the lattice.

Given a lattice, then, we are looking for an ordered link equivalence
that is consistent with the lattice order and is also a total (linear)
order, i.e.\ for any two $e_1, e_2 \in E$, $[e_1] \preceq [e_2]$ or
$[e_2] \preceq [e_1]$.  Many such equivalences exist; for example,
one can always sort the links topologically and assign each link its own
class.  However, such an alignment would be very poor: it would vastly
overestimate the word error between hypotheses.

We initialize the link equivalence such that each initial class consists of
all the links with the same starting and ending times and the same word label.
Starting with this initial partition, the algorithm successively 
merges equivalence classes until a totally ordered equivalence is 
obtained.

Correctness and termination of the algorithm are based on the following
observation. Given a consistent equivalence
relation with two classes $E_1$ and $E_2$ that are not ordered
($E_1 \not\preceq E_2$ and $E_2 \not\preceq E_1$), we can always merge
$E_1$ and $E_2$ to obtain a new equivalence that is still consistent
and has strictly fewer unordered classes (a formal proof is given in
Appendix~\ref{app:proof}).
We are thus guaranteed to create a totally ordered, consistent equivalence
relation after a finite number of steps.

Our clustering algorithm has two stages. We first merge only equivalence classes
corresponding to instances of the same word ({\em intra-word clustering}),
and then start grouping together heterogeneous equivalence classes
({\em inter-word clustering}),
based on the phonetic similarity of the word components.
At the end of the first stage we are able to compute word posterior
probabilities, but it is only after the second stage that we are able to
identify competing word hypotheses.

\subsection{Intra-word clustering}
        \label{sec:intra}

The purpose of this step is to group together all the links corresponding
to same word instance.  Candidates for merging at this step are all the
equivalence classes that are not in relation and correspond to the same word.
The cost function used for intra-word clustering is the following similarity measure between two sets of links:
\newcommand{\SIM}{{\rm SIM}}
\newcommand{\Sim}{{\rm sim}}
\newcommand{\overlap}{{\rm overlap}}
\begin{eqnarray}
\SIM(E_1,E_2) =
  \max\limits_{\begin{array}{l}
                        e_1\in E_1 \\
                        e_2\in E_2
               \end{array}}
                \overlap(e_1,e_2) \cdot p(e_1) \cdot p(e_2)
\label{eq:intra}
\end{eqnarray}
where $\overlap(e_1,e_2)$ is defined as the time overlap between the two
links normalized by the sum of their lengths.  The temporal overlap is 
weighted by the link posteriors so as to make the measure less sensitive
to unlikely word hypotheses.
At each step we compute the similarity between all possible pairs of
equivalence class candidates, and merge those that are most similar.
At the end of this iterative process we obtain a link equivalence relation
that has overlapping instances of the same word clustered together
(although not all such instances necessarily end up in the same equivalence class
due to ordering constraints).

\subsection{Inter-word clustering}
        \label{sec:inter}

In this step we group together equivalence classes corresponding to different
words. Candidates for merging are any two classes that are not in relation.
The algorithm stops when no more candidates are available, i.e.\ a total
order has been achieved.

The cost function used for inter-word clustering is the following similarity
measure based on phonetic similarity between words:
\begin{eqnarray}
\begin{array}{l}
\SIM(F_1,F_2) =
\quad  \operatornamewithlimits{avg}\limits_{
                        \begin{array}{l}
                                w_1 \in \Words(F_1) \\
                                w_2 \in \Words(F_2)
                        \end{array}} 
                        \Sim(w_1,w_2) \cdot p_{F_1}(w_1) \cdot p_{F_2}(w_2)\\
\end{array}
\label{eq:inter}
\end{eqnarray}
where $p_{F}(w) = p(\{e \in F: \Word(e) = w\})$ and $\Sim(\cdot,\cdot)$ is the phonetic similarity between two words,
computed using the most likely phonetic base form.
In our implementation we defined phonetic similarity to be 1 minus the edit
distance of the two phone strings, where phone edit distances are 
normalized by the sum of their lengths.
Other, more sophisticated definitions are conceivable;
for example, the similarity function could be sensitive to the 
phonetic features (e.g.\ vowel/consonant, voiced/unvoiced)
of the phones in the pronunciations.

\subsection{Pruning}
\label{sec:pruning}

\begin{figure}
\begin{center}
\includegraphics[width=0.7\textwidth]{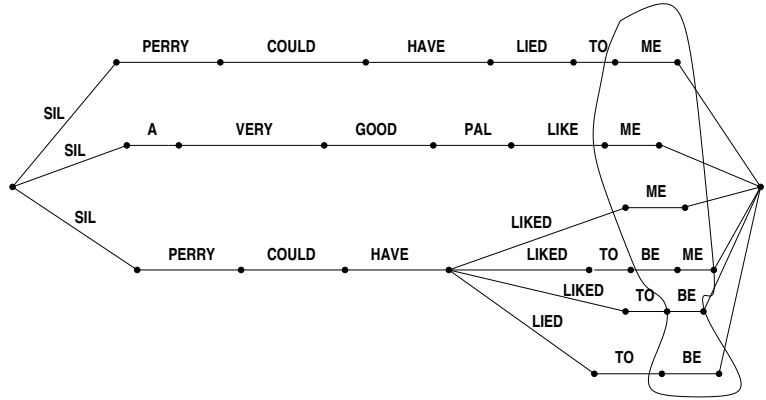}
\end{center}
\caption{Example justifying the pruning step: low probability paths adding order constraints on the words that 
are potential competing candidates.}
\label{problem2}
\end{figure}

Typical word lattices contain links with very low posterior probability.
Such links are negligible in computing the total posterior probabilities
of word hypotheses, but they can have a detrimental effect on the alignment.
This occurs because the alignment preserves consistency with the lattice order,
no matter how low the probability of the links imposing the order.
For example, in Figure~\ref{problem2} we see the words ``BE'' and ``ME'',
which are phonetically
similar and overlap in time, and should therefore be
mutually exclusive.
However, even a single path with ``BE'' preceding ``ME'', no matter how low
in probability, will prevent ``BE'' and ``ME'' from being aligned.

\cbstart
To help eliminate such cases we introduce a preliminary pruning step.
Lattice pruning removes all links whose
posteriors are below an empirically determined threshold.
The equivalence class initialization and subsequent merging only considers
links that survive the initial pruning.
Section~\ref{sec:afterpruning} gives results showing the effect of 
lattice pruning on the overall effectiveness of our algorithm.
Experiments show that word recognition accuracy indeed improves with 
lattice pruning, and that results are not very sensitive to the exact
value of the pruning threshold.
\cbend

\subsection{Confusion networks}
\label{sec:conf_nets}
The total posterior probability of an alignment class can be strictly
less than 1.  That happens when there are paths in the original
lattice that do not contain a word at that position; the missing 
probability mass corresponds precisely to the probability of a deletion
(or null word).
We explicitly represent deletions by a link $e_{-}$ with the corresponding
empty word $\Word(e_{-})= \mbox{``-''}$.

For example, in the lattice in Figure~\ref{fig:example}(a) there are some
hypotheses having ``I'' as the first word, while others have no corresponding
word in that position.  The final alignment thus contains two competing
hypotheses in the first position:
the word ``I'' (with posterior equal to the sum over all hypotheses
starting with that word), and the null word (with posterior equal to the sum
over all other hypotheses).

As illustrated in Figure~\ref{fig:example}(b), the alignment is itself
equivalent to a lattice, which we refer to as a {\em confusion network}.
The confusion network has one node for each equivalence class of original
lattice nodes (plus one initial/final node), and adjacent nodes are 
linked by one edge per word hypothesis (including the null word).
 
We can think of the confusion network as a highly compacted representation
of the original lattice with the property that all word hypotheses are
totally ordered.
As such, the confusion network has other interesting uses besides word
error minimization, some of which will be discussed in
Section~\ref{sec:apl_conf_nets}.

\subsection{The consensus hypothesis} 

Once we have a complete alignment it is straightforward to extract the
hypothesis with the lowest expected word error.
Let $C_i, i = 1, \ldots ,L$ be the final link equivalence classes making
up the alignment.
We need to choose a hypothesis $W = w_1 \ldots w_L$ such that
$w_i = \mbox{``-''}$, or 
$w_i = \Word(e_i)$ for some $e_i \in C_i$.
It is easy to see that the expected word error of $W$ is the sum total
of word errors for each position in the alignment.
Specifically, the expected word error at position $i$ is
\begin{equation}
\begin{array}{lcl}
1 - \sum_{e \in C_i, \Word(e) = w_i} p(e) & \mbox{if} & w_i \neq \mbox{``-''} \\
1 - \sum_{e \in C_i} p(e) & \mbox{if} & w_i = \mbox{``-''} \\
\end{array}
\end{equation}

In other words, the best hypothesis is obtained by picking the 
links in the confusion graph that have the highest posterior probability
among all links at a given position.
This is equivalent to finding the path through the confusion graph with the
highest combined link weight.

\subsection{Score scaling}
        \label{sec:scaling}

Posterior probability estimates are based on a combination of recognizer
acoustic and language model scores [Equation~(\ref{eq:posterior})].
In practice it is necessary to adjust the dynamic ranges of the 
two scores relative to each other to obtain satisfactory results.
This is usually accomplished by multiplying the logarithm of 
the language model score by a constant, the {\em language model weight}.
The absolute scale of the scores does not matter in MAP decoding since 
a maximization is performed.

In computing posterior word probabilities or expected word error, however,
the choice of scale for log scores becomes crucial.  Specifically,
multiplying scores on the log scale controls the peakedness of the 
posterior probability distribution and greatly affects the additive 
combination of posteriors.  An overly peaked posterior concentrates
all its mass on the MAP word hypothesis, causing the consensus or center
hypothesis to converge to that MAP hypothesis.

For these reasons, we need to optimize the scaling of the recognizer scores 
empirically.  We did so using data that was independent of the
test sets used in later experiments, and found the best scaling constant
to be unity for the language model log probabilities, and the
inverse of the language model weight for the acoustic log likelihoods.
We thus arrive at the following expression for the posterior estimate:
\begin{eqnarray}
        \log P(W|A) & = & \log P(W) + \log P(Q | W) + \frac{1}{\lambda} \log P(A | W,Q) - C
\end{eqnarray}
where $\lambda$ is the language model weight,
$P(Q|W)$ is the aggregate pronunciation probability,
and $C$ is a normalization constant that makes $P(W|A)$ sum to unity over 
all hypotheses $W$.\footnote{The constant $C$ corresponds to the 
logarithm of the denominator in Equation~(\ref{eq:posterior}), except for the
fact that the former equation did not include pronunciation probabilities for
simplicity.
We did not find a benefit from using a word insertion penalty for
hypothesis selection, but if a recognition system uses such a parameter
it needs to be adjusted to be compatible with the {\em unscaled}
language model score $\log P(W)$.}

What is noteworthy about this finding is that it presents empirical support
for a commonly given rationalization for the need of the language model weight
in current recognizers.  According to that explanation, the language model
weight compensates for the frame-independence assumption in 
HMM-based acoustic models, which underestimate the joint likelihoods of
correlated, nearby acoustic observations.  It follows that if it 
were not for the invariance to constant factors, the correct 
language model weight is close to 1, whereas the acoustic log likelihoods
should be scaled with a weight less than 1.
Our experiments show that this is indeed the proper weighting.

If we do not use the aggregate pronunciation model $P(Q | W)$ in the
posterior probability computation we favor the words with multiple
baseforms. In our method we add the posterior probabilities of all the
links corresponding to a word in a confusion set regardless of the
pronunciation variant hypothesized for that particular link. Because
\cbstart
all pronunciation variants for a word have probability 1 in the recognizer,
and because words with multiple baseforms tend to occur
more frequently, we need to adjust the likelihood computation so as to
penalize these words.
The simplest way to do this is to apply a uniform pronunciation model in which
all pronunciation variants of a word receive the same probability and sum to
1.  This is the model we employed in all our experiments, though
more elaborate, nonuniform or even context-dependent pronunciation
probabilities are certainly possible.
\cbend

\cbstart
\section{Word recognition experiments}
        \label{sec:results}


We now report experiments investigating the effort of 
the new lattice-based word error minimization algorithm on
word recognition accuracy.
We experimented with two widely used, very different sets of data:
the Switchboard conversational telephone speech corpus \citep*{SWBD}
and the Broadcast News corpus of radio and television news programs
\citep*{Graff:97}.
\cbend

\subsection{Results on the Switchboard corpus}
	\label{sec:results-swb}

\cbstart
The system that generated the lattices used in our experiments was built
using HTK~\citep*{HTK}. It is an HMM-based state-clustered, cross-word
triphone system, trained with about 60 hours of Switchboard speech. The
system uses about 6700 HMM state clusters, each of which refers to a mixture
of 12 Gaussian densities as the state output distribution.
The front end uses MF-PLP derived coefficients with
cepstral means removed per conversation side.
The language model is a backoff trigram model trained
on about 2\.2 million words of Switchboard transcripts.
\cbend

The first set of experiments compares the consensus hypothesis with the sentence
MAP baseline hypothesis. We also report experiments comparing the lattice-based
approach to the $N$-best word error minimization approach.

\subsubsection{Comparison to MAP approach}
\label{sec:compareMap}

\cbstart
The first column in Table~\ref{tab:results} (Set~I) shows results
on a test set used in the 1997 Johns Hopkins
University LVCSR Workshop (WS97) \citep*{WS97}.
This test set consists of 2427 utterances from 19 conversations comprising
about 18\,000 words.

\begin{table}
\caption{Comparison of consensus hypotheses and baseline MAP hypotheses
on two Switchboard test sets.}
\label{tab:results}
\begin{center}
\begin{tabular}{lcc}
\hline
                & \multicolumn{2}{c}{WER (\%)} \\
	\cline{2-3}
Hypothesis      & Set I & Set II \\
\hline
{MAP}& 38\.5 & 42\.9\\
{Consensus}& 37\.3 & 41\.6 \\
$\Delta$ WER &  -1\.2 & -1\.3 \\
\hline
\end{tabular}
\end{center}
\end{table}

Two parameters of our algorithm were optimized on a separate set of
Switchboard lattices that was disjoint from those used in the
recognition test.  The threshold for pruning the original lattices
(cf.\ Section~\ref{sec:pruning}) was set such that word hypotheses with
posteriors less than $10^{-3}$ were eliminated.
Section~\ref{sec:afterpruning} examines how variations of this threshold
affected the results.
Another optimized parameter was the scale of the log posterior probabilities
(cf.\ Section~\ref{sec:scaling}).
As mentioned earlier, we found that posterior scaling with the inverse
of the language model weight (12 in our recognizer) gave best results.

The consensus hypothesis results in an absolute WER reduction of
1\.2\%\footnote{The scoring software used throughout this paper is the
one provided by NIST.} over the baseline, the standard MAP approach.
This difference is statistically significant at the 0\.0001
level.\footnote{The significance test used throughout this paper is a
matched-pairs sign test.}
To verify the consistency of the improvement we ran a
similar experiment on another set of lattices.  Set~II consists of
lattices corresponding to a different set of utterances,
generated using the same acoustic models as used previously.
The baseline WER on this set was more than 4\% higher than that of Set~I.
Set~II results showed a very similar WER reduction over the baseline
(1\.3\%).
\cbend


\subsubsection{Comparison with $N$-best list approach}
 \label{sec:compareNbest}

We also compared the lattice-based consensus hypothesis to the
$N$-best based center hypothesis (cf.~Section~\ref{sec:nbest}).
The maximum number of hypotheses per utterance was 300. We found that 
increasing the number of $N$-best hypotheses to 1000 did not give significant 
error reductions, and therefore concluded that $N=300$ was a safe cutoff
for this experiment.
Table~\ref{tab:center} shows the WER results for the two methods on the
sets of lattices from the previous experiment.

\begin{table}
\caption{Comparison of $N$-best (center) and lattice-based (consensus)
        word error minimization on two Switchboard test sets.}
\label{tab:center}
\begin{center}
\begin{tabular}{lcc}
\hline

                & \multicolumn{2}{c}{WER (\%)} \\
	\cline{2-3}
Hypothesis      & Set I & Set II \\
\hline
MAP& 38\.5 &  42\.9 \\
$N$-best (Center)& 37\.9  & 42\.3 \\
Lattice (Consensus)& 37\.3 & 41\.6\\
\hline
\end{tabular}
\end{center}
\end{table}

We see that there are significant\footnote{The significance level was
0\.001 on Set~I and 0\.0001 on Set~II.} differences between the two
methods, the lattice-based approach being consistently better. For
example, on Set~I the $N$-best center hypothesis results in 0\.6\% improvement
over the baseline, whereas the lattice consensus hypothesis doubles the gain.

\cbstart
Table~\ref{tab:SER} shows that under the new decoding scheme the sentence error rate (SER) in fact increases.
This is to be expected given that, unlike for MAP hypothesis selection,
the objective function for the center and the consensus hypothesis is word 
error, rather than sentence error.
\cbend

\begin{table}
\caption{Word error rate (WER) and sentence error rate (SER) results
on Switchboard test set~I.}
\label{tab:SER}	
\begin{center}
\begin{tabular}{lcc}
\hline
Hypothesis   & WER & SER \\
\hline
{MAP}& 38\.5 & 65\.3\\
$N$-best (Center) & 37\.9 & 65\.6\\
Lattice (Consensus) & 37\.3 & 65\.8 \\
\hline
\end{tabular}
\end{center}
\end{table}


\cbstart
\subsection{Results on the Broadcast News corpus}

We also ran experiments on a set of 1280 lattices corresponding to the 1996
DARPA Hub-4 development test set, drawn from the Broadcast News corpus.
We did not reoptimize the posterior probability scaling or the link
pruning threshold for this experiment.
Instead, we applied what we had learned in the Switchboard experiments: we set the value for the scale to be the language model weight
(which is 13 in this system) and used the pruning value optimized for
Switchboard.

\begin{table}
\caption{WER on Broadcast News showing the breakdown by 
focus condition.  The conditions are F0 (clean read speech),
F1 (conversational speech), F2 (telephone speech),
F3 (speech with background music), F4 (noisy speech),
F5 (non-native speech), and FX (all other conditions).}
\label{tab:BNresults_foci}
  \begin{center}
    \begin{tabular}{lcccccccc}
	\hline
      Hypothesis &  F0  &  F1  &  F2  &  F3  &  F4  &  F5  &  FX  & Overall\\
	\hline
      MAP
      & 13\.0 & 30\.8 & 42\.1 & 31\.0 & 22\.8 & 52\.3 & 53\.9 & 33\.1\\
      $N$-best (Center)
      & 13\.0 & 30\.6 & 42\.1 & 31\.1 & 22\.6 & 52\.4 & 53\.9 & 33\.0\\
      Lattice (Consensus)
      & 11\.9 & 30\.5 & 42\.1 & 30\.7 & 22\.3 & 51\.8 & 52\.7 & 32\.5\\
	\hline
     \end{tabular}
   \end{center}
\end{table}

Overall WER results are presented in the last column of 
Table~\ref{tab:BNresults_foci}. We see that on this corpus the $N$-best approach 
results in almost no improvement, whereas the lattice-based approach 
significantly\footnote{The result is significant at the 0\.005 level.}
reduces the WER (0\.6\% absolute, 1\.8\% relative).

The Broadcast News corpus classifies speech into several different 
{\em focus conditions}, corresponding to different acoustic conditions
and speaking styles.   These are labeled as F1 through FX in 
Table~\ref{tab:BNresults_foci}.
The consensus hypothesis results in improvements in
accuracy across almost all conditions (with no change in F2).  
While there is no obvious correlation between the nature of the focus
conditions and the magnitude of the WER differences,
we observed that the largest improvements (more than
1\% absolute) were obtained on the conditions with lowest and highest 
WER (F0 and FX, respectively).

One significant difference between the Broadcast News corpus and the 
Switchboard data is that the former contains much longer utterances on
average.\footnote{The mean utterance length in the Broadcast News test set
is about 17 words, and about 8 words in the Switchboard test set.}
Longer utterances can be expected to show more of a difference 
between sentence error and word error minimization (in the extreme case
of a one-word sentence, both measures are identical).
This suggests the following diagnostic experiment:
we divided the test utterances into two sets, one containing ``long''
utterances (number of words $> 25$) and one containing ``short'' utterances
(all others), and
measured the WER reduction for the consensus hypothesis on each.
As shown in Table~\ref{tab:shortlong},
we found indeed that the larger gains over the MAP hypothesis
come from the longer utterances.
The absolute WER reduction was 0\.7\% on the
long utterances and only 0\.3\% on the short utterances.

\begin{table}
\caption{Results on Broadcast News by utterance length.}
  \label{tab:shortlong}
\begin{center}
\begin{tabular}{cccc}
\hline
                & \multicolumn{3}{c}{WER (\%)} \\
		\cline{2-4}
Hypothesis      & Short utt. & Long utt. & Total \\
\hline
{MAP}&{33\.3}&{31\.5}&{33\.1}\\
Lattice (Consensus)& {33\.0} & {30\.8} & {32\.5}\\ 
\hline
\end{tabular}
\end{center}
\end{table}

The longer utterances in Broadcast News also provide a likely explanation
for the poor performance of the $N$-best algorithm on this test set.
The sentence hypothesis space grows roughly exponentially with the 
length of an utterance, so given $N$-best lists of fixed length, any method 
based on $N$-best processing will be ignoring a larger and larger portion
of the true posterior distribution of hypotheses.
Lattices, on the other hand, by their combinatorial nature, should capture
a fraction of the true posterior distribution that is roughly constant.
\cbend


\cbstart
\section{Detailed analyses}
\label{sec:analysis}

Here we report several diagnostic experiments and associated
empirical analyses of our algorithm.
All analyses here and in the following section are based
on the WS97 Switchboard test set
(Set~I from Section~\ref{sec:compareMap}).

\subsection{Multiple alignment word error and true word error}

In Section~\ref{sec:wemin} we showed that the key to our lattice-based
word error 
minimization approach was approximating the word error (WE) between two
hypotheses with a 
new string distance MWE, computed on the multiple alignment of the
entire set of hypotheses.
A diagnostic experiment was designed to quantify the difference between
MWE and WE.
We sampled a large number of pairs of hypotheses from the posterior 
distribution represented by our lattices, and compared $\WE$ and $\MWE$
for each pair.
The total number of errors per utterance
(sum of substitutions, deletions, and insertions)
under the two types of alignment differed by only 0\.15 on average.
This suggests that the suboptimal nature of the alignment is a small factor
in practice, and more than justified by the computational 
advantages it affords.
\cbend

\cbstart

\subsection{Effect of lattice pruning}
\label{sec:afterpruning}

As explained in Section~\ref{sec:pruning}, removing low-probability
word hypotheses from the original lattices can improve the quality of
the multiple alignment of hypotheses.  Because poor alignments distort
the MWE metric away from the true WE metric, pruning can actually 
improve the accuracy of the consensus hypothesis.
\cbend

\begin{figure}
\begin{center}
\includegraphics[width=0.5\textwidth,angle=-90]{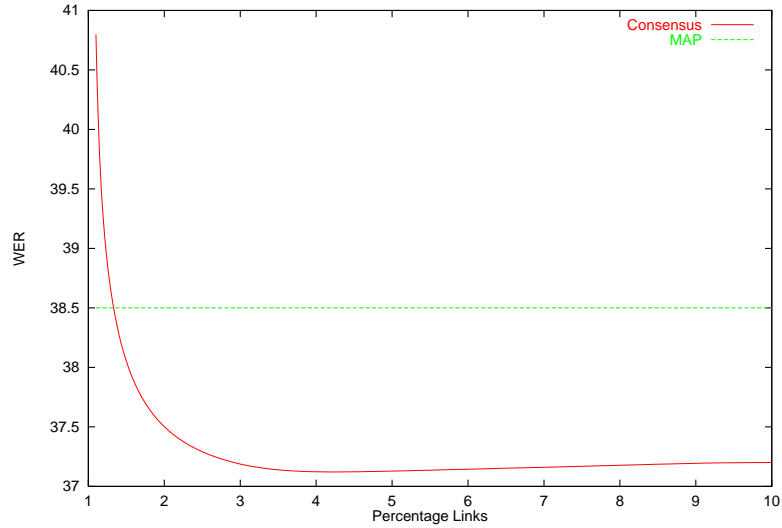}
\end{center}
\caption{Switchboard WER of the consensus hypothesis
when varying the percentage of
links remaining after the pruning step.
The horizontal line marks the baseline WER of the MAP approach.}
\label{fig:consensus_pruning}
\end{figure}

Figure~\ref{fig:consensus_pruning} plots the WER of the consensus
hypothesis as a function of percentage of links retained from the original
lattice.
The original lattices have an average link density of 1350 and
an average node density\footnote{The average link/node density is the
ratio between the total number of links/nodes in the lattice and the
number of words in the reference transcription.} of 370. The lattice
WER, i.e.\ the best WER that can be achieved by choosing a path in the
lattice, is 9\.5\%.  We observe that with less than 2\% of the links
in the original lattice we obtain a hypothesis with the same accuracy
as the baseline MAP hypothesis, and with only 5\% of the links in the
original lattice we obtain a hypothesis that results in 1\.2\%
improvement over the baseline (see Table~\ref{tab:results}).

\subsection{Lattices vs.\ $N$-best lists}
\label{sec:latnbest}

\cbstart
The WER reductions obtained with lattices compared to $N$-best lists
can be attributed to two factors: exploring more hypotheses in the search
for the
minimizer and using more evidence when computing the expected word error.
The contributions of these two factors are quantified in the 
following diagnostic experiment.

$N$-best sentence hypotheses were generated as described in
Section~\ref{sec:compareNbest}.
Also, posteriors for individual word hypotheses were computed from
the corresponding lattices as usual,
using the consensus method.  However, instead of allowing any hypothesis 
from the confusion network, we limited the choice of the best
sentence hypothesis to those in the $N$-best list.
Thus, this result has the benefit of the improved word posterior estimates
based on lattices, but not of the expanded hypothesis selection space.

\begin{table}
\caption{Comparison of $N$-best (center), $N$-best (consensus),
and lattice-based (consensus) word error minimization on the Switchboard
corpus.}
\label{tab:ncons}
\begin{center}
\begin{tabular}{lcc}
\hline
Hypothesis & WER (\%)\\
\hline
MAP& 38\.5 \\
$N$-best (Center)& 37\.9 \\
$N$-best (Consensus)& 37\.6 \\
Lattice (Consensus)& 37\.3 \\
\hline
\end{tabular}
\end{center}
\end{table}

The diagnostic result is shown as ``$N$-best (Consensus)'' in
Table~\ref{tab:ncons}.
Comparing the results for the baseline $N$-best approach, the lattice consensus
approach, and the diagnostic experiment, we conclude that about half of 
the overall WER reduction in the full lattice approach comes from 
improved word posterior estimates, leaving the other half due to
greater freedom in hypothesis selection. 
\cbend

\subsection{Alternative clustering metrics}
	\label{sec:metrics}

\cbstart
In Section~\ref{sec:algorithm} we introduced two similarity metrics
[cf.\ Equations~(\ref{eq:intra}) and~(\ref{eq:inter})]
that form the basis of the clustering procedure.
Here we examine some variations on these metrics.
\cbend

\subsubsection{The importance of time information}

\cbstart
If the input lattice does not have information about the start and end
times of word hypotheses,%
\footnote{In some systems, lattices contain only word identities and 
likelihoods; such lattices are typically used as restricted
language models in multipass speech recognizers.}
we must eliminate
the time overlap term in the similarity metric (\ref{eq:intra}) for the
intra-word clustering stage (Section~\ref{sec:intra}).
In other words, we compute the similarity
between two clusters based solely on the posterior probabilities.  We
ran experiments using this modified metric and found just a slight
increase in WER (0\.15\% absolute).

An alternative is to estimate word time using the available information.
For example, we can compute for each lattice node the length of the 
longest path from the initial node. The path lengths can be determined
either in terms of the number of words or, more accurately, the number of
phones, both of which we expect to be reasonably correlated with actual times.
The $\overlap(.,.)$ term in Equation~(\ref{eq:intra}) is then computed based on
these approximate time marks.
We carried out experiments using word times estimated from phone counts,
and obtained exactly
the same WER as in the original experiments with time information.
\cbend

\subsubsection{The importance of phonetic similarity}

\cbstart
We now investigate the importance of phonetic similarity in the
inter-word clustering stage (Section~\ref{sec:inter}).
We ran experiments with a modified similarity function based entirely
on word posterior probabilities, i.e.\ we removed the $\Sim(.,.)$ term from
Equation~(\ref{eq:inter}):
\begin{equation}
\begin{array}{l}
\SIM(F_1,F_2) =
\quad  \operatornamewithlimits{avg}\limits_{
                        \begin{array}{l}
                                w_1 \in \Words(F_1) \\
                                w_2 \in \Words(F_2)
                        \end{array}} 
                        p_{F_1}(w_1) \cdot p_{F_2}(w_2)\\
\end{array}
\label{eq:noph}
\end{equation}
We were surprised to find no change in WER as a result of this
change,
and conclude that the topology of the lattice alone constrains
the alignment process sufficiently, as long as equivalence classes
with high posterior probability are merged first.

We might still want to use phonetic similarity in clustering if
our goal is to post-process confusion networks.
For example, while low probability hypotheses do not seem to affect 
the quality of the best hypothesis as defined by our present algorithm,
phonetic similarity might improve the alignment of hypotheses with low
probability. This in turn might be important once we move to more
sophisticated methods for choosing among the words of one equivalence class,
as suggested in Section~\ref{sec:apl_conf_nets}.

Furthermore, we found that phonetic similarity does become
important when the lattices do not contain time information.
Table~\ref{tab:notime} shows results where time information
was not used during intra-word clustering (cf. the previous section).
We see that phonetic similarity does improve the accuracy of the
consensus hypotheses by about 0\.2\% absolute in this case.
This suggests that time information and phonetic similarity are
somewhat complementary for the purpose of word alignment.
\cbend

\begin{table}
\caption{Effect of phonetic similarity when no time information is used in
clustering (Switchboard test set~I).}
\label{tab:notime}
\begin{center}
\begin{tabular}{lc}
\hline
Method &  WER (\%) \\
\hline
MAP & 38\.5 \\
Consensus (no times, with phonetic similarity)& 37\.3 \\
Consensus (no times, w/o phonetic similarity) & 37\.5 \\
\hline
\end{tabular}
\end{center}
\end{table}

\subsubsection{The role of the posterior probabilities}

The clustering procedure is a greedy algorithm. Whenever we merge two
clusters, we add new constraints to the partial order. Consequently,
some equivalence classes that could have been merged earlier are
no longer candidates for merging.
\cbstart
For this reason it is very important to have a robust
similarity measure that does the right thing first and foremost on the
high probability words.
Here we investigate the importance of posterior
probabilities as weights in computing cluster similarities.
\cbend

\begin{figure}
\begin{center}
\includegraphics[width=0.7\textwidth]{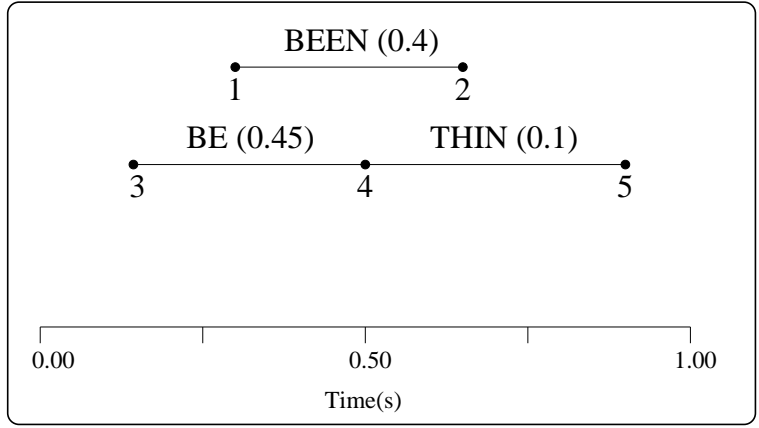}
\end{center}
\caption{The role of posterior probabilities in clustering}
\label{fig:problem4}
\end{figure}

Figure~\ref{fig:problem4} shows one situation that was
encountered quite often in our experiments. In this example we have to
choose between merging ``BE'' and ``BEEN'' or ``BE'' and
``THIN''. If ``BE'' and ``BEEN'' are the actual competitors and we
prefer to merge ``BEEN'' and ``THIN'', then we might end up with a
consensus hypothesis in which both ``BEEN'' and ``BE'' are deleted
because the deletion has a high posterior in both classes.
However, the fact that both ``BE'' and ``BEEN'' have high
posteriors, and at the same time there is no path containing both of
them, suggests precisely the fact that they are candidates for the same
word.

\begin{table}
\caption{WER of the consensus hypothesis when no posterior probabilities
are used in clustering (Switchboard test set~I).}
\label{tab:usepost}
\begin{center}
\begin{tabular}{ccccc}
\hline
Method      &  WER (\%) & Substitutions &  Deletions & Insertions\\
\hline
{Similarity with posteriors}&{37\.3}&{4365}&{1837}&{547} \\
{Similarity w/o posteriors}& {37\.6} & {4143} & {2165} & {494}\\ 
\hline
\end{tabular}
\end{center}
\end{table}

\cbstart
The above scenario is consistent with the results we obtained
when we experimented without posterior weighting in the similarity metrics.
As shown in Table~\ref{tab:usepost}, we observed a large increase in
the number of deletion errors, and a moderate increase in the overall
error rate.
\cbend

\section{Confusion networks revisited}
\label{sec:apl_conf_nets}

\cbstart
Our method produces two main outputs:
the consensus hypothesis and the confusion network.
\cbend
Up to this point we have focused on the properties
of the consensus hypothesis.
Now, we study the properties of confusion networks and show
how they can be used for tasks other than word error minimization.

\subsection{Correct hypothesis rank}

Confusion networks can serve as a compact representation of the
hypothesis space for post-processing speech recognizer output with
other knowledge sources.  
A relevant question for such applications is how word hypotheses ranked
by the confusion network compare to the truth.

\begin{figure}
\begin{center}
\includegraphics[width=0.6\textwidth]{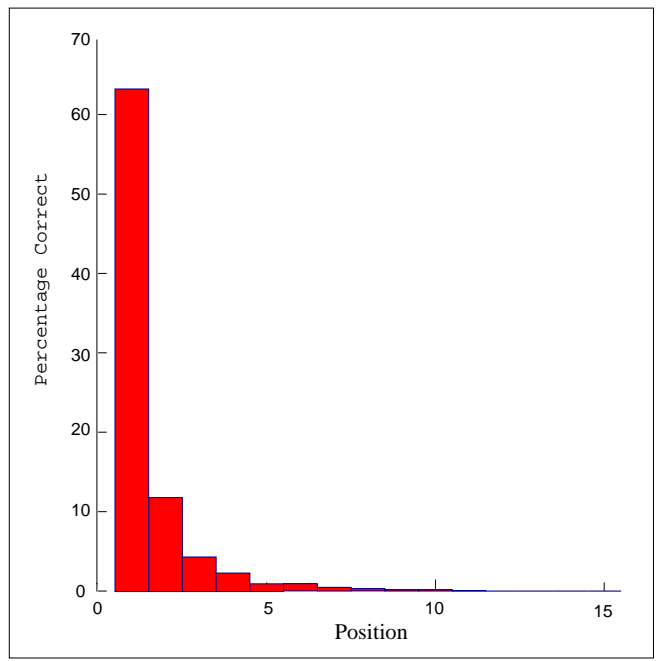}
\end{center}
\caption{Histogram of the rank of the correct word in the confusion networks
built for Switchboard test set~I.}
\label{fig:histo}
\end{figure}

Words in each confusion set in the confusion network can be ranked based on their
posterior probabilities. If we align the reference transcription to the 
network and compute the rank of the correct word in each confusion set we obtain
the histogram in Figure~\ref{fig:histo}.
We see that very rarely is the 
rank of the correct word greater than 10, and rarely greater than 7.
This suggests limiting the confusion set size before using the confusion networks for
some other applications.
The results also show that if we find a method for better discriminating
between the best two candidates in each confusion set, we can improve the recognizer's
accuracy by 10\%. This suggests that there is space for potential improvement over the current choice,
but we should note that the task of finding the right discriminator
is, of course, not trivial. 

Analyzing the properties of the confusion sets further we extracted some other interesting statistics:
\begin{itemize}
\item When a confusion set has only one candidate, this candidate is correct 97\% of the time; 25\% of the confusion sets have this property.
\item When a confusion set has only two candidates, the highest-scoring candidate is correct 90\% of the time; 25\% of the confusion sets have this property.
\end{itemize}
This means that in 50\% of the cases we can predict the correct word with high accuracy. These words could therefore be used as features for disambiguating 
the rest of the confusion sets. 
We also found that in confusion sets containing the best two candidates with close scores,
choosing the one with the highest score is almost as good as picking randomly.
This suggests that even little additional information could help the
disambiguation procedure. 

In summary, these statistics suggest that confusion networks are an excellent
representation if one aims to narrow down the large search space associated
with speech recognition to a small set of mutually exclusive, single-word
candidates.  By constructing confusion networks one can start to
apply machine learning techniques based on discrete k-way classifiers 
to enhance the quality of the recognition hypothesis, for example,
by incorporating more sophisticated linguistic knowledge.

\cbstart

\subsection{Hypothesis space size}

\begin{figure}
\begin{center}
\includegraphics[width=0.6\textwidth,angle=-90]{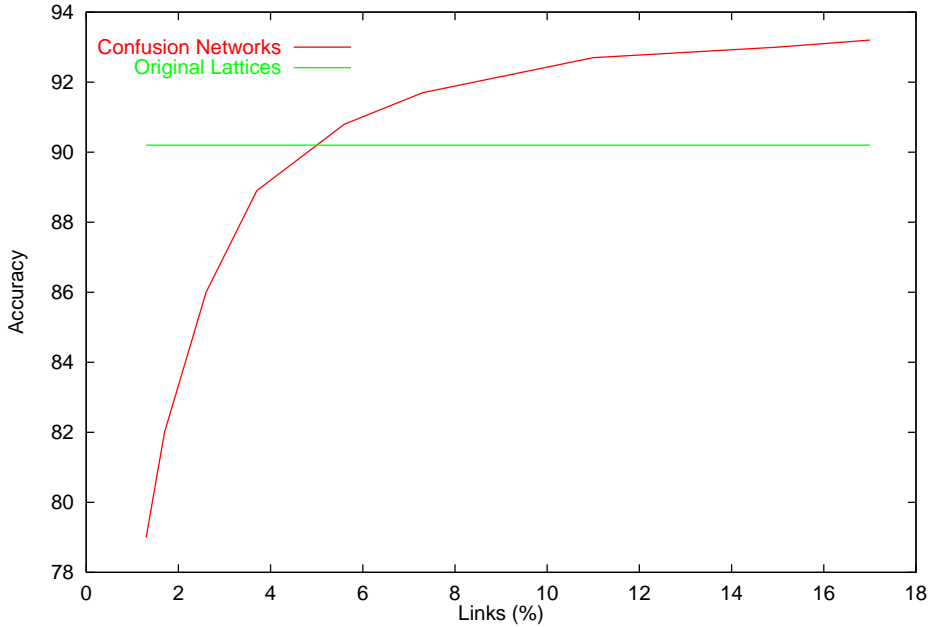}
\end{center}
\caption{Confusion network accuracy on Switchboard test set~I
as a function of the percentage of
links in the original lattice remaining after the pruning step.}
\label{fig:conf_pruning}
\end{figure}

In Section~\ref{sec:afterpruning} we discussed the effect of lattice 
pruning (as defined in Section~\ref{sec:pruning})
on the accuracy of the consensus hypothesis.
We now ask a related question: How does lattice pruning affect the
hypothesis space represented in the confusion network?
To investigate the issue we first look at the
{\em accuracy of confusion networks}, defined as for general lattices,
i.e.\ 1 minus the WER of the best path through the graph. 
Figure~\ref{fig:conf_pruning} shows the relationship between lattice pruning
and the word accuracy of confusion networks.
As before, we quantify the degree of pruning by 
the percentage of links retained from the original lattices.

As shown, with only 5\% of the links in the original lattice  we obtain
word graphs with the same accuracy.  Given that 5\% of the links in the
original lattices represent only 20\% of the word types, this suggests
that only a small percentage of the word hypotheses in the original
lattice are relevant to further processing.
Figure~\ref{fig:conf_pruning} also shows that if we retain more than
5\% of the original links we obtain networks with even better accuracy
than the original lattices. This is a consequence of the fact that the
confusion network connects high-probability words that were
disconnected in the original, partially ordered representation.

One might suspect that despite the comparatively small number of 
{\em word} hypotheses, pruned confusion networks still contain a
larger number of {\em sentence} hypotheses, since all word hypotheses 
may be combined (except those in the same confusion set).
This might present a problem for post-processing algorithms, such 
as a parser, that are concerned with sentence-level analyses.

We explicitly computed the number of paths
in both our original WS97 lattice set and the pruned confusion networks.
For clustering purposes, the lattices were pruned to 5\% of the original links,
so that the resulting
confusion networks had an accuracy equal to that of the original full lattices.
We found that the number of total paths in the 2427 original lattices was
$10^{81}$, compared to $10^{80}$ paths in the pruned confusion networks.
Furthermore, in 90\% of the utterances the confusion network
contained fewer paths than the original lattice.

We conclude that, based on the number of hypotheses alone, confusion networks
should not be less well suited to sentence-level post-processing than
recognition lattices.
Of course it might still be the case that other properties of the 
sentence hypotheses allowed by confusion network cause problems in
certain natural language processing applications.
In the next section we examine a different pruning scheme that 
guarantees that the final set of sentence hypotheses is a proper subset
of the set of original sentence hypotheses.
\cbend

\subsection{Consensus-based lattice pruning}

Word lattices are typically used as intermediate representations of
the hypothesis space prior to further recognition passes or more 
sophisticated hypothesis evaluation techniques.
Two conflicting goals for lattice generation are therefore compactness
(to save computation in further processing) and accuracy (to preserve
the correct hypothesis as often as possible).
The standard method to control the tradeoff between lattice size and
accuracy is likelihood-based pruning: paths whose overall
score differs by more than a threshold from the best-scoring path are
removed from the word graph.

Several techniques have been developed explicitly to reduce
the size of lattice representations while preserving all hypotheses in 
the lattice, e.g.\ by determinizing and minimizing the lattice 
as a finite state network \citep*{Mohri:97} or by node merging in
nondeterministic lattices \citep*{Weng:darpa98}.
In the previous section we saw that confusion networks can serve 
as a compacted lattice representation that can also {\em improve}
lattice accuracy by adding plausible paths.

\cbstart
Here we propose an alternative pruning approach that uses confusion
networks merely as a filter on the hypothesis space,
i.e.\ without adding paths to the original lattice.
\cbend
In this approach we first construct a confusion
network, prune it as discussed previously, and then {\em intersect}
the original lattice with the pruned confusion network.
In other words, we preserve only original lattice paths that are also
in the pruned confusion network.
Final lattice size is controlled by the pruning threshold as applied to
the confusion network.
The effect of this {\em consensus-based pruning} is that paths are preserved,
even if they have overall low probability, if they consist of pieces that
have individually high posteriors.

\begin{figure}
\begin{center}
\includegraphics[width=0.6\textwidth,angle=-90]{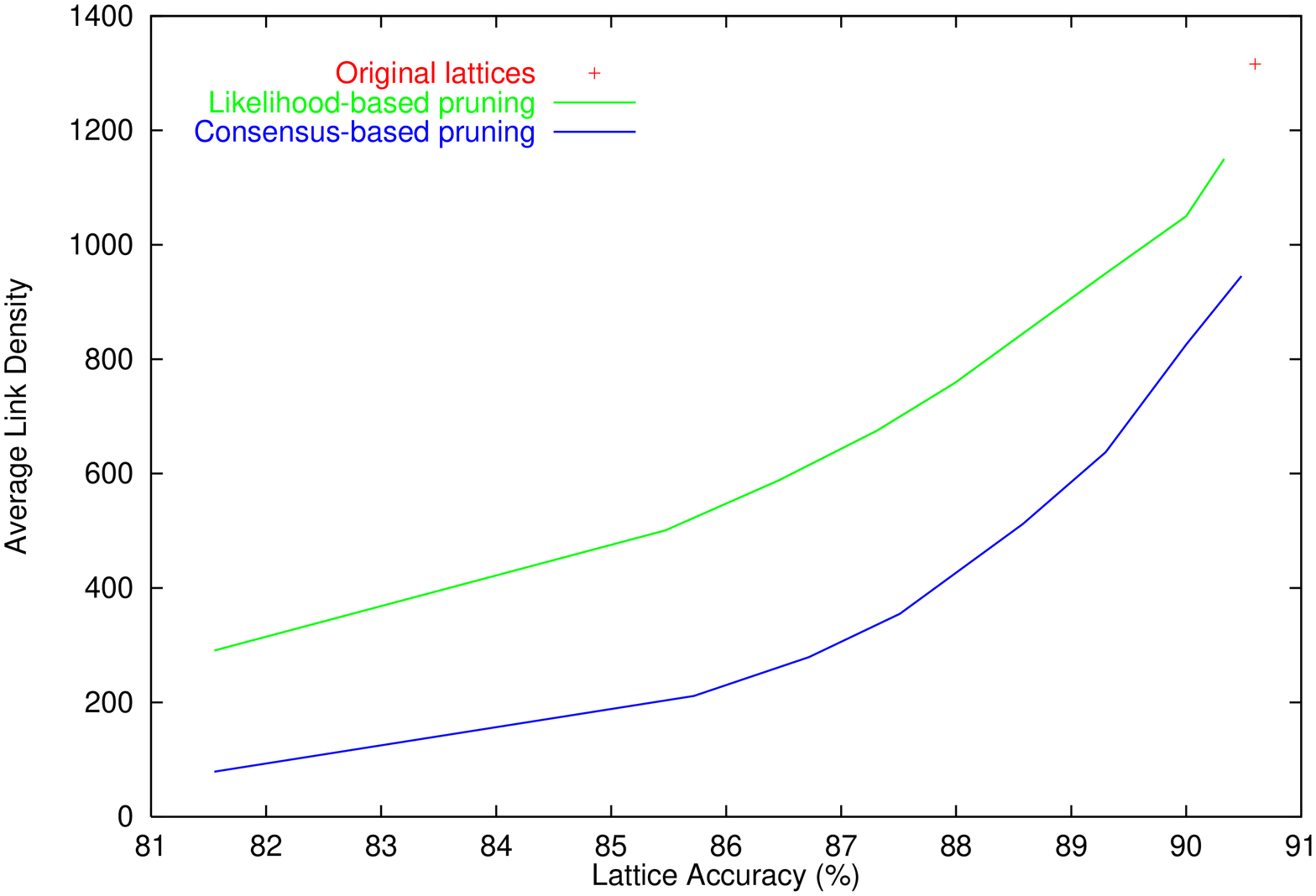} \\
\includegraphics[width=0.6\textwidth,angle=-90]{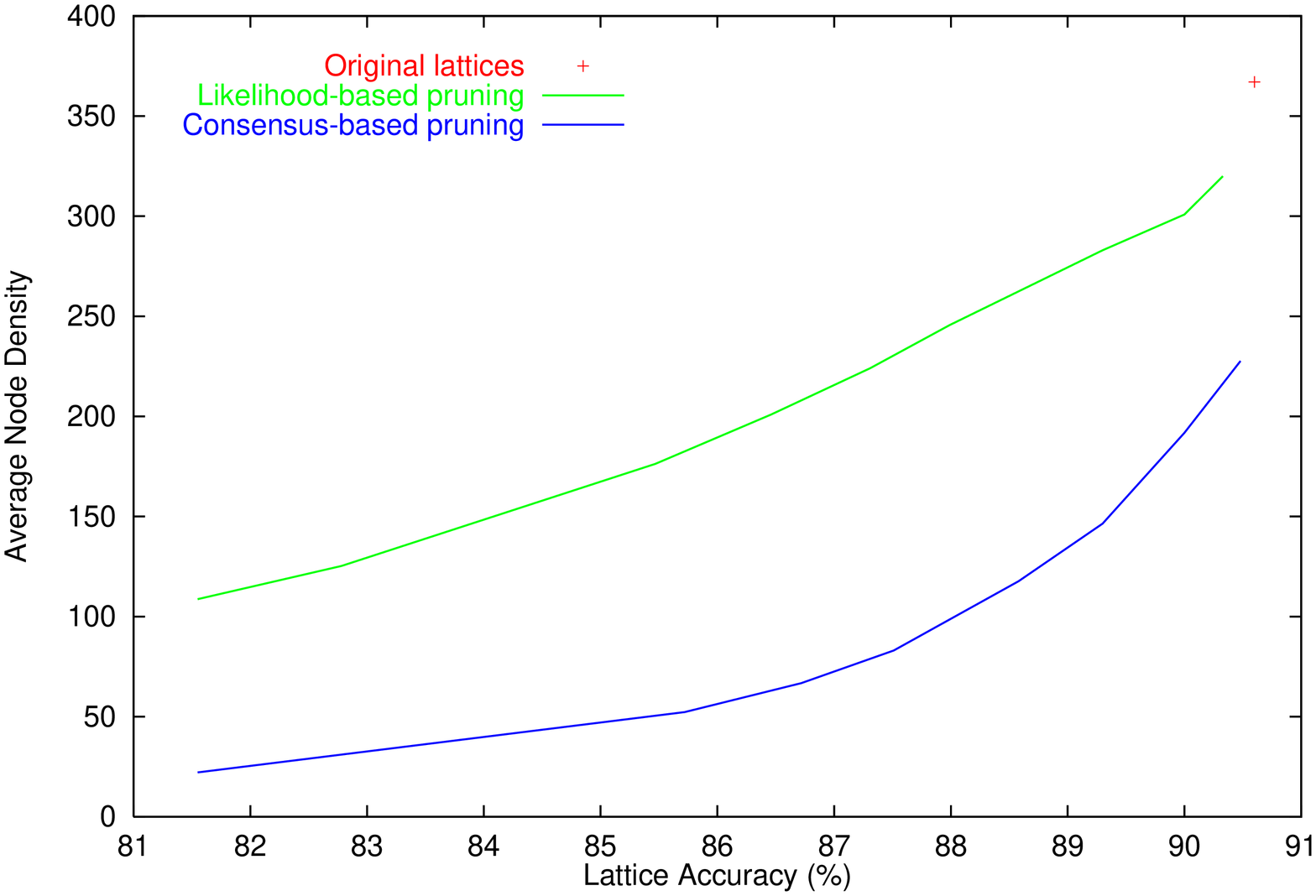}
\end{center}
\caption{Effectiveness of consensus-based pruning in terms of average
node and link density reduction at different lattice accuracy levels
when compared with the standard likelihood-based pruning. The original
lattices had an average link density of 1350, an average node density of
370, and an oracle accuracy of 90\.5\%.}
\label{fig:lattice_pruning}
\end{figure}

\cbstart
Figure~\ref{fig:lattice_pruning} shows the size of lattices (in terms of 
node and link densities) as a function of lattice accuracy,
for both the traditional likelihood-based pruning and consensus-based pruning.
We see that
consensus-pruned lattices are three to four times smaller at equivalent
accuracy levels.
We can conclude that while consensus-based pruning involves the extra
processing step of confusion network construction and filtering, this effort
could be worthwhile when a good size-accuracy tradeoff is desired for
expensive post-processing algorithms.
\cbend

\subsection{Other applications and related work}

\subsubsection{Confidence annotation}

The word-level posterior probabilities included in the
confusion network provide valuable information for several
tasks that require assessing the validity of individual word recognition
hypotheses.
The most obvious such task is confidence annotation of recognition
output, i.e.\ estimating the probability of correctness of each
word.  The posterior probability according to a recognizer's scoring function
(as incorporated in the confusion network) is itself a confidence
measure.
However, the posterior probabilities thus computed tend to overestimate
the word accuracy since they are based on only a subset of the infinite 
word hypothesis space.

\cbstart
A common technique uses an additional estimator that converts
$N$-best or lattice-based word posterior probabilities to unbiased
word correctness probabilities.  Logistic regression \citep*{Siu:97},
decision trees \citep*{Evermann:00}, and 
neural networks \citep*{Weintraub:icassp97} have been used for this purpose.
These techniques also allow miscellaneous other features and knowledge
sources to be incorporated into word confidence estimation.
However, it has been found that the recognizer-based word posteriors 
are usually among the most informative features predicting word correctness
\citep{Siu:97,Weintraub:icassp97}.
Still, confusion networks could provide several additional input features for
confidence estimators, such as the difference in posteriors between candidates,
their posterior entropy, or the number of candidates in each confusion set.

Several papers investigate the usefulness of
lattice-based features for confidence prediction
\citep*{Kemp:97,Wessel:99}.  \citet{Kemp:97} show that the
features obtained from a word graph are the most important predictors,
although in their study the predictive feature considered is the
posterior probability computed for lattice links rather than for words.
\citet{Wessel:99} describe methods to
compute word posterior probabilities by summing over the posterior
probabilities of the links associated with
a common time frame. This purely time-based alignment approach contrasts
with ours, which relies primarily on lattice topology.
\citet{Evermann:00} investigate both time-dependent posteriors and
posteriors based on confusion networks for confidence estimation.
Their results show both approaches to
give comparable estimates, with one or the other at a slight
advantage depending on the acoustic models and lattice sizes used.
\cbend

\subsubsection{Word spotting}

Another application of word posterior estimation is word spotting.
As shown by \citet{Weintraub:95}, high-performance word spotting can be
achieved by estimating word posteriors from the $N$-best output of a
large vocabulary recognizer. In that approach, a time-based criterion
was used to identify word hypotheses across $N$-best lists that correspond
to the same word.  An estimate of the word's probability of occurrence
was then obtained by adding the posterior probabilities for the corresponding
sentence hypotheses. The confusion network algorithm
is a lattice-based generalization of that approach, and we would therefore
expect improved word spotting accuracy based on confusion network
posterior probabilities.

\cbstart
\subsubsection{System combination}

\begin{figure}
\begin{center}
\includegraphics[width=0.7\textwidth]{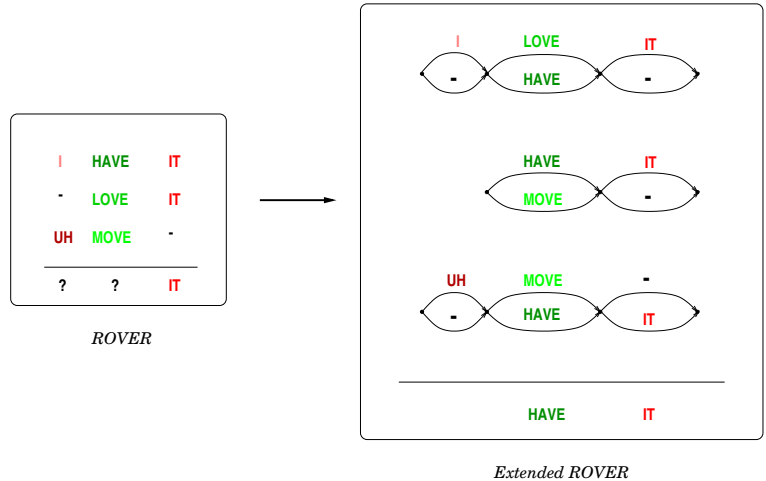}
\end{center} 
\caption{Generalized ROVER algorithm: voting among confusion networks.}
\label{fig:rover}
\end{figure}

\citet*{Fiscus:97} developed a popular algorithm known as {\em ROVER}
(Recognizer Output Voting Error Reduction) for
combining the outputs of several recognizers to yield higher-accuracy
hypotheses.
In the standard ROVER algorithm the 1-best outputs
\cbend
from multiple recognizers are aligned to ``vote'' on a
new hypothesis that is generated by splicing together pieces of 
the original hypotheses.  ROVER is similar to the confusion network
construction algorithm in that it uses multiple alignments and word
confidences for voting, although its task is simpler since
only linear input hypotheses are considered.
Its benefit is also limited by the fact that only a single word
hypothesis is available from each recognizer.
This suggests that an improved ROVER algorithm can be obtained 
by using the full confusion networks and associated posterior probabilities
from each recognizer. For example, if there is a tie in one of the alignment
columns,
it might help to know that one of the words was a close second choice of
some of the other recognizers (see Figure~\ref{fig:rover}).  

\cbstart
The idea of a generalized ROVER algorithm based on alignment of confusion
networks was recently developed and
implemented independently by \citet*{EvermannWoodland:nist2000}.
The technique was applied in the NIST Hub-5 conversational speech recognition
evaluation and found to give a 0\.3\% absolute reduction in WER compared to
the standard ROVER algorithm.  An improvement of the same magnitude was 
found with a different implementation based on combining $N$-best hypotheses lists
\citep{StolckeEtAl:nist2000}.
\cbend

\section{Other lattice-based approaches}
	\label{sec:other}

To make word error minimization on lattices feasible we took the
approach to slightly modify the word error function (to a close
approximation) so as to allow an efficient, exact determination of the
best hypothesis.  An alternative approach recently developed by
\citet{Goel:CSL} takes a different approach and
searches the lattice directly using the original word error cost
function.  To keep the search feasible an A${}^\ast$ heuristic search
is employed, along with various approximations and pruning strategies.
Results show a consistent word error reduction beyond that achieved with
the $N$-best approximation.  
Based on results reported in Section~\ref{sec:results-swb} and by
\citet{Goel:CSL},
the algorithm described here compares favorably with the A${}^\ast$ approach.%
\footnote{Test set I used in our 
experiments and the MAP baseline accuracy coincide with that
used by \citet{Goel:CSL}.}

\cbstart
The two lattice-based approaches to word error minimization each have
distinct advantages.
The A${}^\ast$ method can be generalized to certain loss functions other than
standard word error, in particular to weighted variants of word error.
Practical limits to generalizability arise from the need to compute lower
and upper bounds on the expected loss.
The main strength of our algorithm is that it also 
produces a confusion network representation of the hypothesis space together
with associated word posterior probabilities.
\cbend

\section{Conclusion}
        \label{sec:conclusion}

We  have developed a new method for extracting from a recognition word
lattice hypotheses that minimize expected word error,
unlike the standard MAP scoring approach that minimizes sentence error.
The core of the  method is a clustering procedure
that identifies mutually supporting and competing word hypotheses in a 
lattice, constructing a total order over all word hypotheses.
Together with word posterior probabilities computed from recognizer scores,
this allows an efficient extraction of the hypothesis with a minimum
expected number of errors.
Experiments on two speech corpora show that this approach results
in significant WER reductions and significant lattice compression. 
Our method is also an estimator of word posterior probabilities and as 
such it can be used for other tasks, such as confidence annotation,
word spotting, and system combination.

{\footnotesize

\vspace{\baselineskip}
\noindent
We thank Mitch Weintraub for valuable discussions and suggestions on
the word hypothesis clustering problem, and the anonymous reviewers for
valuable suggestions.  Thanks also go to Frederick Jelinek,
Sanjeev Khudanpur, David Yarowsky, Radu Florian, Ciprian Chelba, and Jun
Wu for useful feedback during STIMULATE meetings.  Dimitra Vergyri
kindly provided the lattices used in the experiments. The work reported
here was supported in part by NSF and DARPA under NSF grant IRI-9618874
(STIMULATE).  The views and conclusions contained in this document
are those of the authors and should not be interpreted as necessarily
representing the official policies, either expressed or implied, of
the funding agencies.
 
\bibliography{all}

} 

\appendix

\small

\newpage
\section{Optimized N-best rescoring algorithm}
        \label{app:nbest-alg}
        
Here we give the pseudo-code for an optimized version of the N-best
word error minimization algorithm, implementing Equation~\ref{eq:center}.
The N-best hypotheses are denoted by $W^{(1)}, \ldots, W^{(N)}$.
$P(\cdot | A)$ are the posterior probabilities estimated as 
described in Section~\ref{sec:scaling}, and $\WE(., .)$ is the 
word error function.
\newcommand{\FOR}{{\bf for}}
\newcommand{\DO}{{\bf do}}
\newcommand{\IF}{{\bf if}}
\newcommand{\WHILE}{{\bf while}}
\newcommand{\THEN}{{\bf then}}
\newcommand{\ELSE}{{\bf else}}
\newcommand{\END}{{\bf end}}
\newcommand{\GOTO}{{\bf goto}}
\begin{quote}
\renewcommand{\baselinestretch}{1.0}\normalsize
\begin{tabbing}
${\it BestHyp} := 0$; \\
${\it BestError} := \infty$; \\
\FOR\ \= $i := 1, \ldots, N$ \DO \+\\
        ${\it ThisError} := 0$; \\
        \FOR\ \= $k := 1, \ldots, N$ \DO \+\\
                ${\it ThisError} := {\it ThisError} 
                                 + P(W^{(k)} | A) WE(W^{(k)}, W^{(i)})$; \\
                \IF\ \= ${\it ThisError} \geq {\it BestError}$ \THEN \+\\
\<\<\<(*)\>\>\>
                        \GOTO\ next; \-\\
                \END \-\\
        \END \\

        \IF\ \= ${\it ThisError} < {\it BestError}$ \THEN \+\\
                ${\it BestError} := {\it ThisError}$; \\
                ${\it BestHyp} := i$; \-\\
        \END \\
        next: \-\\
\END \\
/*  {\it BestHyp} contains index of best hypothesis */
\end{tabbing}
\end{quote}
The test and loop exit marked by (*) are crucial for efficiency.
With it, the worst-case complexity of the algorithm is still $O(N^2)$;
however, in practice the inner loop is exited after a few iterations
whose average number is independent of $N$.
On Switchboard N-best lists we measured runtimes (excluding constant overhead)
that scale perfectly linearly with $N$.
\cbstart
In practice, the optimized algorithm is quite fast:
The Switchboard 300-best lists were processed in about 
$0.1 \times$ real time using a 400 MHz Pentium-II CPU.
\cbend

\newpage

\section{Lattice alignment algorithm}
        \label{app:lattice-alg}

Here we give a concise description of the algorithm that constructs the 
edge equivalence relation that results in a complete lattice alignment.

As outlined in Section~\ref{sec:algorithm}, the algorithm proceeds in three
stages.
\noindent The initial link equivalence classes are formed by word identity and start and
end times:
\[
L_{w,t_1,t_2} = \{e \in E  | \Word(e) = w, \Itime(e) = t_1, \Ftime(e) = t_2\}
\]
The initial partial order $\preceq$ is given as the transitive closure of 
the edge order $\leq$ defined on the lattice (see Section~\ref{sec:align}).

The next step performs {\em intra-word clustering}, i.e., merging of classes containing the same words:
\begin{quote}
\renewcommand{\baselinestretch}{1.0}\normalsize
\begin{tabbing}
\DO\    \= \+ \\
        ${\it MaxSim} := 0$;\\
        \FOR\ \= all $L_1 \in \mathcal{E}$ \+\\
                \FOR\ \= all $L_2 \in \mathcal{E}$ \+\\
                        \IF\    \= $\Words(L_1) = \Words(L_2)$ and
                                        $L_1 \not\preceq L_2$ and
                                        $L_2 \not\preceq L_1$ \THEN \+\\
                                ${\it sim} := \SIM(L_1,L_2)$;\\
                                \IF\ \= $sim > {\it MaxSim}$ \THEN \+\\
                                        ${\it MaxSim} := {\it sim}$;\\         
                                        ${\it bestSet}_1 := L_1$;\\
                                        ${\it bestSet}_2 := L_2$; \-\\
                                \END \-\\
                        \END \-\\
                \END \-\\
        \END \\

        $L_{\it new} := {\it bestSet}_1 \cup {\it bestSet}_2$; \\

        \FOR\ \= all $L_i \in \mathcal{E}$ \+\\
                \IF\ \= $L_i \preceq {\it bestSet}_1$ or
                         $L_i \preceq {\it bestSet}_2$ \THEN \+\\
                        let $L_i \preceq L_{\it new}$; \-\\
                \END \-\\
        \END \\
        \FOR\ \= all $L_i \in \mathcal{E}$ \+\\
                \IF\ \= ${\it bestSet}_1 \preceq L_i$ or
                         ${\it bestSet}_2 \preceq L_i$ \THEN \+\\
                        let $L_{\it new} \preceq L_i$; \-\\
                \END \-\\
        \END \\
        \FOR\ \= all $L_i \in \mathcal{E}$ \+\\
            \FOR\ \= all $L_j \in \mathcal{E}$ \+\\
                 \IF\ \= ($L_i \preceq {\it bestSet}_1$ and ${\it bestSet}_2 \preceq L_j$) or 
                         ($L_i \preceq {\it bestSet}_2$ and ${\it bestSet}_1 \preceq L_j$)  \THEN \+\\
                        let $L_i \preceq L_j$; \-\\
                \END \-\\
            \END \-\\
        \END \\
        
        $\mathcal{E} := \mathcal{E} \cup \{L_{\it new}\}
                        \setminus \{{\it bestSet}_1, {\it bestSet}_2\}$; \-\\
\WHILE\ ${\it MaxSim} > 0$
\end{tabbing}
\end{quote}
The notation used is described in Section~\ref{sec:algorithm}.
In this stage we use the similarity metric $\SIM(\cdot,\cdot)$
described in Section~\ref{sec:intra}. 
The partial order relation $\preceq$ is updated minimally upon merging of 
equivalence classes so as to keep it consistent with the
previous order; an inductive definition of $\preceq$ is given as part of the
correctness proof in Appendix~\ref{app:proof}.

The final step, {\em inter-word clustering}, uses the same algorithm
as the previous step, but we replace the $\SIM$ metric with one that is
based on phonetic similarity as described in
Section~\ref{sec:inter} and eliminate the condition $\Words(L_1) = \Words(L_2)$
that the words corresponding to two equivalence class candidates should be the same.

\cbstart
The complexity of the entire confusion network construction algorithm
is dominated by the hypothesis alignment, which is of order
$O(T^3)$, where $T$ is the number of links in the word lattice.
As shown in Figure~\ref{fig:consensus_pruning}, we can prune about 95\%
of the links in the original lattices without affecting the accuracy of the
consensus hypothesis.
Hence, $T$ is kept small and runtimes are very reasonable in practice:
on a 400 MHz Pentium-II processor, the Switchboard lattices were processed
in about $0.55 \times$ real time on average.
\cbend

\newpage
\section{Proof of correctness}
\label{app:proof}

To complete a formal proof of the correctness of 
the lattice alignment algorithm we need to show that successive merging
of equivalence classes is possible while maintaining a consistent 
partial order on the classes.
To demonstrate termination of the
algorithm, we need to establish that classes suitable for merging can
be found as long as any two classes are unordered.  (Since the initial
number of classes is finite and is decremented in each merging step, this 
guarantees that the algorithm terminates with a total order of classes
after a finite number of steps.)

Here we give a constructive proof by induction that a consistent 
partial order can be defined for the initial equivalence partition,
and maintained for each merging step.
Recall that a relation $\preceq$ is a {\em partial order} if and only if
it has the following properties:
\begin{description}
\item[\it Reflexivity:]  For all $x$, $x \preceq x$.
\item[\it Transitivity:] For all $x, y, z$, $x \preceq y$ and $y \preceq z$ imply
                $x \preceq z$.
\item[\it Antisymmetry:] For all $x, y$, $x \preceq y$ and $y \preceq x$ imply
                $ x = y$.
\end{description}
The concept of consistency was defined in Section~\ref{sec:align}.

For the first part of the proof, we must show that the relation we define
on the initial set of equivalence classes is a consistent partial order.
We form the initial partition $\mathcal{E}_0$ by word identity and 
start/ending times as described in Appendix~\ref{app:lattice-alg}.
A partial order $\preceq_{0}$ on $\mathcal{E}_{0} = \{ L_{w,t_1,t_2} \}$
is given as follows.
If $\leq$ is the partial order on links introduced in Section~\ref{sec:align}
define the relation $\mathcal{R}$ on 
$\mathcal{E}_{0}$ as follows:
\begin{center}
$E_1\ \mathcal{R}\ E_2$ iff there exists
$e \in E_1$ and $f \in E_2$ such that $e \leq f$.
\end{center}
and define $\preceq_{0}$ as the transitive closure of $\mathcal{R}$. 
Reflexivity of $\mathcal{R}$ holds, since $e \leq e$ for any link $e \in E$.
We claim that $\mathcal{R}$ is an antisymmetric relation.
Let $E_1$ and $E_2$ be two sets of links in $\mathcal{E}_0$ such that $E_1 \neq E_2$, $E_1\ \mathcal{R}\ E_2$ and $E_2\ \mathcal{R}\  E_1$. Consequently there exist $e_{11} \in E_1, e_{21} \in E_2$ with $e_{11} \leq e_{21}$  and  there exist $e_{22} \in E_2, e_{12} \in E_1$ with $e_{22} \leq e_{12}$. Given that all the links in the same equivalence class have the same start and end times, and that if two links in the lattice are $\leq$ ordered then the start and end times are ordered accordingly, we obtain the following:
\[
\Ftime(e_{11}) \leq \Itime(e_{21}) = \Itime(e_{22}) < \Ftime(e_{22}) \leq \Itime(e_{12}) = \Itime(e_{11})
\]
which is a contradiction. Thus, $\mathcal{R}$ is a reflexive and antisymmetric relation, which makes its transitive closure $\preceq_{0}$ a partial order. The consistency of $\preceq_{0}$ comes directly from the definition of $\mathcal{R}$.

In the second part of the proof we have to show that if we start with a consistent partial order and at each step merge only equivalence classes that are not ordered (under the current order), we still have a consistent partial order on the new set of equivalence classes. The proof proceeds by induction over the merging steps. 

Let $\mathcal{E}_i$ be the set of equivalence classes and $\preceq_i$ the consistent partial order on $\mathcal{E}_i$ at step $i$, and $L_1$ and $L_2$  
the two candidate equivalence classes that can be merged at step $i$, hence having the property $(L_1 \not\preceq_i L_2$ and $L_2 \not\preceq_i L_1)$.
We define $L_{\it new} = L_1 \cup L_2$ as the new merged equivalence class.

The set of equivalence classes at step $i+1$ is
\[
\mathcal{E}_{i+1} = \mathcal{E}_{i} \cup \{L_{\it new}\}  \setminus \{L_1, L_2\}
\]
and the relation $\preceq_{i+1}$ is derived from $\preceq_{i}$ as follows:
\begin{itemize}
\item if $U \preceq_{i} V$ then $U \preceq_{i+1} V$ for any $U,V \in \mathcal{E}_{i+1}$, $U \neq L_{\it new}$ and $V \neq L_{\it new}$
\item if $U \preceq_{i} L_1$ or $U \preceq_{i} L_2$ then $U \preceq_{i+1} L_{\it new}$
\item if $L_1 \preceq_{i} V$ or $L_2 \preceq_{i} V$ then $L_{\it new} \preceq_{i+1} V$
\item if $(U \preceq_{i} L_1$ and $L_2 \preceq_{i} V)$ or $(U \preceq_{i} L_2$ and $L_1 \preceq_{i} V)$ then $U \preceq_{i+1} V$
\item $L_{\it new} \preceq_{i+1} L_{\it new}$
\end{itemize}

\noindent We claim that $\preceq_{i+1}$ as defined above is a partial order.
 
\noindent {\bf Reflexivity} holds, because $\preceq_{i}$ is reflexive and $L_{\it new} \preceq_{i+1} L_{\it new}$.\\
{\bf Antisymmetry} can be shown by contradiction. Let us assume that there exist $U, V \in \mathcal{E}_{i+1}$, $U \neq V$ such that $U \preceq_{i+1} V$ and $V \preceq_{i+1} U$.

{\it Case 1:} $U \neq L_{\it new}$ and $V \neq L_{\it new}$ \\
Given that $U \neq V$, the fact that $U \preceq_{i+1} V$ comes either from $U \preceq_{i} V$ or ($U \preceq_{i} L_1$ and $L_2 \preceq_{i} V$) or ($U \preceq_{i} L_2$ and  $L_1 \preceq_{i} V$). The same holds for $V \preceq_{i+1} U$.
We would then have to consider all nine combinations, but the proofs for many of them are very similar, and therefore we spell out only the most interesting cases:
\begin{itemize}
\item If $U \preceq_{i} V$ and $V \preceq_{i} U$ then $U = V$ (antisymmetry of $\preceq_{i}$). Contradiction with the assumption $U \neq V$.
\item If $U \preceq_{i} V$ and $V \preceq_{i} L_1$ and $L_2 \preceq_{i} U$ then $L_2 \preceq_{i} L_1$ (transitivity of $\preceq_{i}$). This contradicts the condition that $L_1$ and $L_2$ be unordered to be candidates for merging.
\item If $U \preceq_{i} L_1$ and $L_2 \preceq_{i} V$ and $V \preceq_{i} U$ then $L_2 \preceq_{i} L_1$. Same contradiction as in the previous case.
\item If $U \preceq_{i} L_1$ and $L_2 \preceq_{i} V$ and $V \preceq_{i} L_2$ and $L_1 \preceq_{i} U$ then $U = L_1$ and $V = L_2$ (antisymmetry of $\preceq_{i}$). Contradiction with $U, V \in \mathcal{E}_{i+1}$.
\end{itemize}

{\it Case 2:} $U = L_{\it new}$\\
By definition, $L_{\it new} \preceq_{i+1} V$ only if either $L_1 \preceq_{i} V$ or $L_2 \preceq_{i} V$, and similarly for $V \preceq_{i+1} L_{\it new}$. Therefore we have to consider four cases, but only two of them are structurally distinct:
\begin{itemize}
\item If $L_1 \preceq_{i} V$ and $V \preceq_{i} L_1$ then  $V = L_1$ (antisymmetry of $\preceq_{i}$). Contradiction with $V \in \mathcal{E}_{i+1}$.
\item If $L_1 \preceq_{i} V$ and $V \preceq_{i} L_2$ then  $L_1 \preceq_{i} L_2$ (transitivity of $\preceq_{i}$). Contradiction with the condition that $L_1$ and $L_2$ be unordered.
\end{itemize}

{\it Case 3:} $V = L_{\it new}$ is similar to Case~2.
 
\noindent {\bf Transitivity}  Let us consider $U, V, W \in \mathcal{E}_{i+1}$ such that $U \preceq_{i+1} V$ and $V \preceq_{i+1} W$. We have to show that 
$U \preceq_{i+1} W$.

{\it Case 1:} $U \neq L_{\it new}$, $V \neq L_{\it new}$ and $W \neq L_{\it new}$\\
Using the previous remark concerning the reasons for $U \preceq_{i+1} V$ and $V \preceq_{i+1} W$, we have the following cases:
\begin{itemize}
\item If $U \preceq_{i} V$ and $V \preceq_{i} W$, then $U \preceq_{i} W$ (transitivity of $\preceq_{i}$). Consequently, $U \preceq_{i+1} W$.
\item If $U \preceq_{i} V$ and $V \preceq_{i} L_1$ and $L_2 \preceq_{i} W$, then $U \preceq_{i} L_1$ (transitivity of $\preceq_{i}$) and $L_2 \preceq_{i} W$. Thus, $U \preceq_{i+1} W$ (definition of $\preceq_{i+1}$).
\item If $U \preceq_{i} L_1$ and $L_2 \preceq_{i} V$ and $V \preceq_{i} W$, then $U \preceq_{i} L_1$ and $L_2 \preceq_{i} W$. Consequently, $U \preceq_{i+1} W$.
\end{itemize}
It is not possible to have $U \preceq_{i} L_1$ and $L_2 \preceq_{i} V$ and $V \preceq_{i} L_1$ and $L_2 \preceq_{i} W$ because this would imply $L_2 \preceq_{i} L_1$.
Also, if we had $U \preceq_{i} L_1$ and $L_2 \preceq_{i} V$ and $V \preceq_{i} L_2$ and $L_1 \preceq_{i} W$ we could infer $L_2 = V$, which contradicts 
$V \in \mathcal{E}_{i+1}$.

{\it Case 2:} $U = L_{\it new}$\\
The only possible combinations are
\begin{itemize}
\item $L_1 \preceq_{i} V$ and $V \preceq_{i} W$,  in which case $L_1 \preceq_{i} W$, therefore $U \preceq_{i} W$ (definition of $\preceq_{i+1}$)
\item $L_2 \preceq_{i} V$ and $V \preceq_{i} W$,  in which case $L_2 \preceq_{i} W$, therefore $U \preceq_{i} W$ (definition of $\preceq_{i+1}$)
\end{itemize}
The other combinations are not possible. For example, if $L_1 \preceq_{i} V$ and $V \preceq_{i} L_1$ and $L_2 \preceq_{i} W$ then $V = L_1$ which 
contradicts $V \in \mathcal{E}_{i+1}$.

{\it Case 3:} $W = L_{\it new}$ is similar to Case~2.

{\it Case 4:} $V = L_{\it new}$. 
Given that $U \preceq_{i+1} L_{\it new}$ only if $U \preceq_{i} L_1$ or $U \preceq_{i} L_2$, we have the following cases:
\begin{itemize}
\item If $U \preceq_{i} L_1$ and $L_1 \preceq_{i} W$ then $U \preceq_{i} W$ (transitivity of $\preceq_{i}$), and so $U \preceq_{i+1} W$.
\item If $U \preceq_{i} L_2$ and $L_1 \preceq_{i} W$ then $U \preceq_{i} W$ (definition of $\preceq_{i+1}$), hence $U \preceq_{i+1} W$.
\end{itemize}

\noindent We have shown that the new relation is a partial order. We still need to show that it is consistent, i.e., if $e \leq f$ then $[e]_{i+1} \preceq_{i+1} [f]_{i+1}$. 
\begin{itemize}
\item If $[e]_{i+1} \neq L_{\it new}$ and $[f]_{i+1} \neq L_{\it new}$ then $[e]_{i+1} = [e]_{i}$ and $[f]_{i+1} = [f]_{i}$. Given that $[e]_{i} \preceq_{i} [f]_{i}$ (consistency of $\preceq_{i}$) we obtain  $[e]_{i+1} \preceq_{i+1} [f]_{i+1}$ (definition of $\preceq_{i+1}$).
\item If $[e]_{i+1} = L_{\it new}$ (which implies $[f]_{i+1} \neq L_{\it new}$) then $e \in L_1$ or $e \in L_2$, and so $L_1 \preceq_{i} [f]_{i}$ or $L_2 \preceq_{i} [f]_{i}$ (consistency of $\preceq_{i}$). Consequently, $L_{\it new} \preceq_{i+1} [f]_{i}$. Given that $[f]_{i} = [f]_{i+1}$ we obtain $[e]_{i+1} \preceq_{i+1} [f]_{i+1}$.
\item Same for $[f]_{i+1} = L_{\it new}$
\end{itemize}

\end{document}